\pdfoutput=1
\documentclass[twoside,leqno,twocolumn]{article}

\usepackage[letterpaper]{geometry}

\usepackage{ltexpprt}
\usepackage{hyperref}
\usepackage{graphicx}
\usepackage{microtype}      
\usepackage{graphicx}
\usepackage{amsmath}
\usepackage{caption}
\usepackage{amssymb}
\usepackage{subcaption}
\usepackage{xcolor} 
\usepackage{wrapfig}
\usepackage{lipsum}
\usepackage{algorithm}
\usepackage[algo2e]{algorithm2e} 
\usepackage{mathrsfs} 
\usepackage{wrapfig}
\usepackage{esvect}
\usepackage{algpseudocode}
\newcommand{\methodName}{\texttt{MedDiff}}
\newcommand{\bx}{\mathbf{x}}
\newcommand{\bz}{\mathbf{z}}
\newcommand{\bI}{\mathbf{I}}

\newcommand{\sigsq}{\sigma_{i}^{2}}

\newcommand{\sequence}{\bx_1,\bx_2,\dots,\bx_T}
\newcommand{\bal}{\bar{\alpha}}

\newcommand{\stdnormal}{\mathcal{N}(\mathbf{0},\mathbf{I})}
\newcommand{\Exp}{\mathbb{E}}
\newcommand{\Pini}{p_{\theta}(\bx_{0:T})}
\newcommand{\Pinter}{p_{\theta}(\bx_{t-1}|\bx_{t})}
\newcommand{\Qini}{q(\bx_{1:T}|\bx_0)}
\newcommand{\Qinter}{q(\bx_{t}|\bx_{t-1})}
\newcommand{\loglikelihood}{\log p_{\theta}(\bx_{0})}
\newcommand{\bep}{\mathbf{\epsilon}}
\newcommand{\pred}[1]{\frac{\bx_{#1} - \sqrt{1-\bal_{#1}}\bep_{\theta}(\bx_{#1},#1)}{\sqrt{\bal_{#1}}}}

\definecolor{purple}{rgb}{1,0,1}
\newcount\Comments  
\newcommand{\kibitz}[2]{\ifnum\Comments=1\textcolor{#1}{#2}\fi}

\begin{document}

\title{\Large MedDiff: Generating Electronic Health Records using Accelerated Denoising Diffusion Model}
\author{Huan He\thanks{huan\_he@hms.harvard.edu, Harvard University}
\and Shifan Zhao\thanks{Emory University}
\and Yuanzhe Xi$^\dagger$
\and Joyce Ho$^\dagger$}

\date{}

\maketitle


\fancyfoot[R]{\scriptsize{Copyright \textcopyright\ 20XX by SIAM\\
Unauthorized reproduction of this article is prohibited}}





\begin{abstract} \small\baselineskip=9pt Due to patient privacy protection concerns, machine learning research in healthcare has been undeniably slower and limited than in other application domains. High-quality, realistic, synthetic electronic health records (EHRs) can be leveraged to accelerate methodological developments for research purposes while mitigating privacy concerns associated with data sharing. The current state-of-the-art model for synthetic EHR generation is generative adversarial networks, which are notoriously difficult to train and can suffer from mode collapse. Denoising Diffusion Probabilistic Models, a class of generative models inspired by statistical thermodynamics, have recently been shown to generate high-quality synthetic samples in certain domains. It is unknown whether these can generalize to generation of large-scale, high-dimensional EHRs. In this paper, we present a novel generative model based on diffusion models that is the first successful application on electronic health records. Our model proposes a mechanism to perform class-conditional sampling to preserve label information. We also introduce a new sampling strategy to accelerate the inference speed. We empirically show that our model outperforms existing state-of-the-art synthetic EHR generation methods.\end{abstract}

\section{Introduction}
Recent digitisation of health records has enabled the training of deep learning models for precision medicine, personalised prediction of risks and health trajectories \cite{Rajkomar2018ScalableAA,Miotto2016DeepPA}. However, there are still patient privacy concerns that need to be accounted for in order to aggregate more data to train more robust models. As such, it is still hard for researchers to obtain access to real electronic health records (EHRs). One approach to mitigate privacy risks is through the practice of de-identification such as perturbation and randomization \cite{ElEmamh1139,McLachlan2016UsingTC}. However, this approach is vulnerable to re-identification \cite{4531148}.
Another approach is to use synthetic datasets that capture as many of the complexities of the original data set (e.g., distributions, non-linear relationships, and noise). Synthetic EHRs can yield a database that is beyond de-identification and hence are immune to re-identification. 
Thus, generating realistic, but not real data is a key element to advance machine learning for the healthcare community.
There have been several distinguished efforts conducted in a variety of domains about synthetic data (EHR) generation \cite{bing2022conditional, DBLP:journals/corr/abs-2012-10020, pmlr-v68-choi17a, Torfi2020CorGANCC,yan2020generating, he2021age, cai2022autm}.
Unfortunately, existing proposed algorithms predominantly adopt a variant of Generative Adversarial Network (GAN) \cite{Gan,he2022gdaam}, auto-encoder \cite{AE}, or a combination of both. While GANs and autoencoders are natural and widely used candidates for generation, there are several noticeable drawbacks of these models, including mode-collapse for GANs or poor sample diversity and quality for autoencoders. 

\begin{figure*}
    \centering
    \includegraphics[ width=0.7\linewidth]{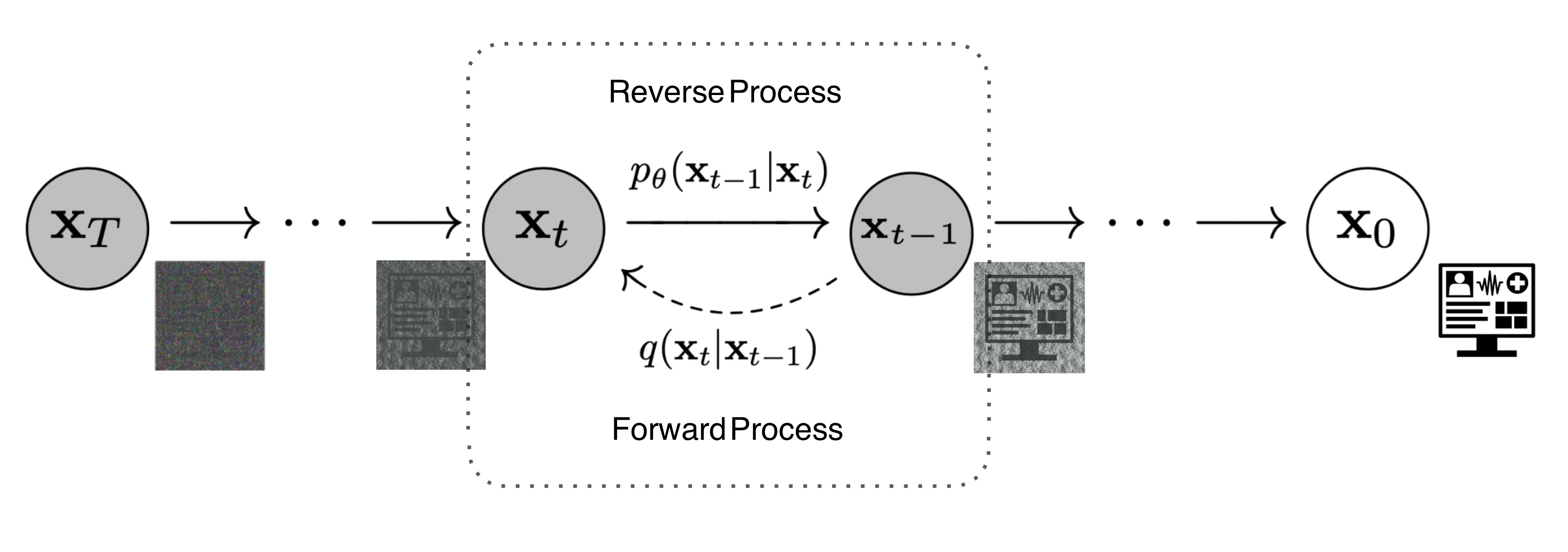} 
    \caption{The idea of diffusion model for generating synthetic EHRs. The forward process adds noise to the original patient record while the reverse process generates a synthetic sample by removing noise. }
    \label{fig:medDIFF}
\end{figure*}

In recent years, diffusion (score) based models \cite{ho2020denoising, denoise,score} have emerged as a family of powerful generative models that can yield state-of-the-art performance across a range of domains, including image and speech synthesis \cite{chen2021wavegrad,vahdat2021scorebased}. 
Diffusion models have been shown to achieve high-quality, diverse samples that are superior to their GAN-based counterparts.
Other key advantages of diffusion models include ease of training and tractability, in contrast to GANs, and speed of generation, in contrast to autoregressive models \cite{pmlr-v32-gregor14}. This leads to a natural question: Is a diffusion model promising for generating synthetic EHRs as well? We answer this in the affirmative, by introducing \methodName, a novel denoising diffusion probabilistic model. \methodName, shown in Figure \ref{fig:medDIFF}, generates high quality, robust samples while also being simple enough for practitioners to train.
We further accelerate the generation process of \methodName~using Anderson acceleration \cite{anderson1965iterative}, a numerical method that can improve convergence speed of fixed-point sequences.
In summary, our contributions are as follows:
\begin{itemize}
    \item We investigate the effectiveness of diffusion based models on generating discrete EHRs.
    \item We propose to accelerate the generation process, a main drawback of diffusion models. 
    \item We introduce a novel conditioned sampling technique to generate discriminative synthetic EHRs.
    \item We show that \methodName~can generate realistic synthetic data that mimics the real data and provides similar predictive value.
\end{itemize}

\section{Related Work \& Background}
To the best of our knowledge, \methodName~is the first work to leverage the idea of diffusion based modeling to generate EHRs. Since our work casts insight on the effectiveness of generating EHR via diffusion based models, we leave the generation of more complex EHRs (e.g., multiple data sources, temporal data) for future investigation.
Here, we discuss the related works on synthetic EHRs generation and related diffusion based models. 

\subsection{Synthetic EHR generation}
Closely related to this work are recent efforts that leverage deep generative models for synthesizing EHRs \cite{doi:10.1161/CIRCOUTCOMES.118.005122,pmlr-v68-choi17a,Torfi2020CorGANCC}.
MedGAN \cite{pmlr-v68-choi17a} and CorGAN \cite{Torfi2020CorGANCC} were introduced to generate patient feature matrices.
However, these works rely heavily on the performance of a pre-trained autoencoder model to reduce the dimensionality of the latent variable.
Without the pre-trained autoencoder, these GAN-based models can fail to generate high-quality samples, highlighting the difficulty of using these models when generalizing to multiple institutions (e.g., smaller clinics or different patient distributions). \methodName~builds on diffusion models and does not require a pre-trained encoder.

\subsection{Diffusion Models}
First proposed in \cite{denoise}, diffusion models are a family of latent variable generative models characterized by a forward and a reverse Markov process.
The forward process gradually adds noise to the original data sample, whereas the reverse process undoes the gradual noising process.
In the reverse process, the sampling starts with the $T$th noise level, $x_T$, and each timestep produces less-noisy samples, $x_{T-1}, x_{T-2}, ..., x_0$. 
In essence, the diffusion model learns the ``denoised" version from $x_{t-1}$ to $x_t$.

Diffusion models have several advantages over existing generative modeling families.
They do not rely on adversarial training which can be susceptible to mode collapse and are difficult to train.
They also offer better diversity coverage and can accommodate flexible model architectures to learn any arbitrary complex data distributions.
With respect to image and speech synthesis, diffusion-based models can achieve high-quality, diverse samples that are superior to their GAN counterparts.

Ho et al. \cite{ho2020denoising} proposed a specific parameterization of the diffusion model to simplify the training process using a score matching-like loss that minimizes the mean-squared error between the true noise and the predicted noise.
They also note that the sampling process can be interpreted as equivalent to Langevian dynamics, which allows them to relate the proposed denoising diffusion probabilistic models (DDPM) to score-based methods in \cite{score}.
Denoising diffusion implicit models (DDIM) was recently proposed to offer a more flexible framework based on DDPM \cite{song2021denoising}.
Here, we introduce the forward process, the reverse process, training, and sampling of DDIM.

\subsection{DDIM}

\subsubsection{Forward Process}
Given a data distribution $\bx_0 \sim q(\bx_0)$, diffusion models gradually add noise to the original data distribution until it loses all the original information and becomes an entirely noisy distribution (as shown by the $x_T$ sample in Figure \ref{fig:medDIFF}).
DDPM convolves $q(\bx_0)$ with an isotropic Gaussian noise $\mathcal{N}(0,\sigsq \bI)$ in T steps to produce a noise corrupted sequence $\sequence$. $\bx_T$ will converge to an isotropic Gaussian distribution as $T\rightarrow \infty$. 
However, for DDIM, a different variance schedule is used to produce $\sequence$. The explicit input  distribution $q$ of DDIM is derived as :
\begin{equation}
    q(\bx_{1:T}|\bx_0) := q(\bx_{T}|\bx_0)\prod_{t=2}^{T}q(\bx_{t-1}|\bx_{t},\bx_0)
\end{equation}
where $q(\bx_{T}|\bx_0) = \mathcal{N}(\sqrt{\bal}\bx_0, (1-\bal_T)\bI)$ and for all $t>1$,
\begin{align}
    q(\bx_{t-1}|\bx_{t},\bx_0) =& \mathcal{N} (\sqrt{\bal_{t-1}}\bx_0 +  \\ 
    &\sqrt{1-\bal_{t-1}-\sigma^2}\frac{\bx_t - \sqrt{\bal_t}\bx_0}{\sqrt{1-\bal_{t}}}, \sigma^2\bI) \nonumber
\end{align}
Here $\bal$ controls the scale of noise added at each time step. At the beginning, noise should be small so that it is possible for the model to learn well, e.g., $\bal_1>\bal_2\dots>\bal_T$.

The DDIM distribution is parameterized to guarantee the marginal density is equivalent to DDPM. However, the key difference between DDPM and DDIM is that the forward process of DDIM is no longer a Markov process. This allows acceleration of the generative process as multiple steps can be taken. Moreover, different reverse samplers can be utilized by changing the variance of the reverse noise. This means DDIM can be compatible with other samplers.

\subsubsection{Backward process}
For DDPM, the forward process is defined by a Markov chain, and thus the true sample can be recreated from Gaussian noise $\bx_T \sim\stdnormal $ by reversing the forward process. However, as noted above, DDIM relies on a family of non-Markovian processes. Denote $p_{\theta}$ as a parameterized neural network, the reverse process with a prior $p_{\theta}(\bx_{T}) = \mathcal{N}(\mathbf{0},\bI)$ can be computed as 
\begin{equation}
\begin{split}
      p_{\theta}(\bx_{0}|\bx_1) &= \mathcal{N}\Big(\pred{1},\sigma_{1}^2\bI\Big)  \\
      p_{\theta}(\bx_{t-1}|\bx_t)& = q(\bx_{t-1}|\bx_{t},\pred{t}),\quad t>1
\end{split}
\end{equation}

\subsection{Training}
The training process of DDPM and DDIM is based on optimizing the variational lower bound on the negative log likelihood:
\begin{align}
     \Exp[-\loglikelihood] &\le \Exp_{q}\Big[-\log \frac{\Pini}{\Qini}\Big] \\
     &= \Exp_{q}\Big[-\log p(\bx_{T}) - \nonumber \\
     & ~~~~~\sum_{t\geq 1}\log \frac{\Pinter}{\Qinter}\Big] =: L \nonumber 
\end{align}
As shown in Ho et al. \cite{ho2020denoising}, this is equivalent to the following loss function:
\begin{equation}
    \Exp_{\bx_{0},\bep}\Big[ \frac{\beta_t^2}{2\sigma_{t}^2\alpha_{t}(1-\bal)}\|\bep - \bep_{\theta}(\sqrt{\bal}\bx_0 + \sqrt{1-\bal}\bep,t)\|\Big]
\end{equation}

\subsection{Sampling}
After training, sampling can be done in DDIM using the following equation:
\begin{align}
    \bx_{t-1} =& \sqrt{\bal_{t-1}}\Big(\pred{t}\Big) +\\
    & \sqrt{1-\bal_{t-1} -\sigma_{t}^2}\bep_{\theta}(\bx_t, t) + \sigma_t \bz_t, \nonumber 
\end{align}
where $\bz_{t} \sim \stdnormal$.
If we denote $\frac{\sqrt{1-\bal_s}}{\sqrt{\bal_s}}$ by $\lambda_{s}$, then the updating rule for continuous time is
\begin{equation}
    \bx_{s} = \frac{\lambda_{s}}{\lambda_{t}}[\bx_{t} - \bal_{t}\bep_{\theta}(\bx_t)] + \bal_{s}\bep_{\theta}(\bx_t)
    \label{eqn:update}
\end{equation}
To generate high-quality samples, it requires repeating the above updating rule (denoising process) multiple times and thus make the sampling procedure much slower than other generative models that only need one pass. 
A recent paper \cite{salimans2022progressive} showed that DDIM is an integrator of the probability flow ordinary differential equation (ODE) defined in \cite{song2020score}:
\begin{equation}
\label{eq:integrator-ode}
    d\bx = [f(\bx,t) - \frac{1}{2}g^{2}(t)\nabla_{\bx}\log p_{t}(\bx)]dt
\end{equation}
where $d\bx = f(\bx,t)dt + g(t)dW$ is a stochastic differential equation and $W$ is Brownian motion. In practice $\nabla_{\bx}\log p_{t}(\bx) = \frac{\bal_t \bep_{\theta}(\bx_t) - \bx_t}{\lambda_{t}^2}$ so $\bx_{s}$ can be regarded as the integrator of $\frac{1}{2}[\bal_t \bep_{\theta}(\bx_t) - \bal_{t}^2\bx_{t}]$.


\section{MedDiff}
In this section, we introduce the three components of \methodName. We first introduce the base architecture. Next, we propose a numerical method to accelerate the generation process. We then equip \methodName~with the ability to conduct conditioned sampling.  

\subsection{Base Architecture}
In our preliminary experiments, we observed that a simple, fully connected diffusion model is sufficient for low-dimensional data. However, applying DDPM developed for images and audio yields unsatisfying results. 
The current best architectures for image diffusion models are U-Nets \cite{ronneberger2015unet,ho2020denoising}, which are a natural choice to map corrupted data to reverse process parameters. However, they are tailored for 2D signals such as images and video frames. Yet, this may not be a viable option in numerous applications over 1D signals especially when the training data is scarce or application specific. As a result, \methodName~uses a modified U-Net architecture including larger model depth/width, positional embeddings, residual blocks for up/downsampling, and residual connection re-scale. Additionally, we use the 1d convolutional form of U-Net to generate a vector for each patient. Under this base architecture, \methodName~can better capture the neighboring feature correlations.

\begin{figure*}
\centering
\begin{subfigure}{.48\textwidth}
    \centering
    \includegraphics[width=1.25\linewidth]{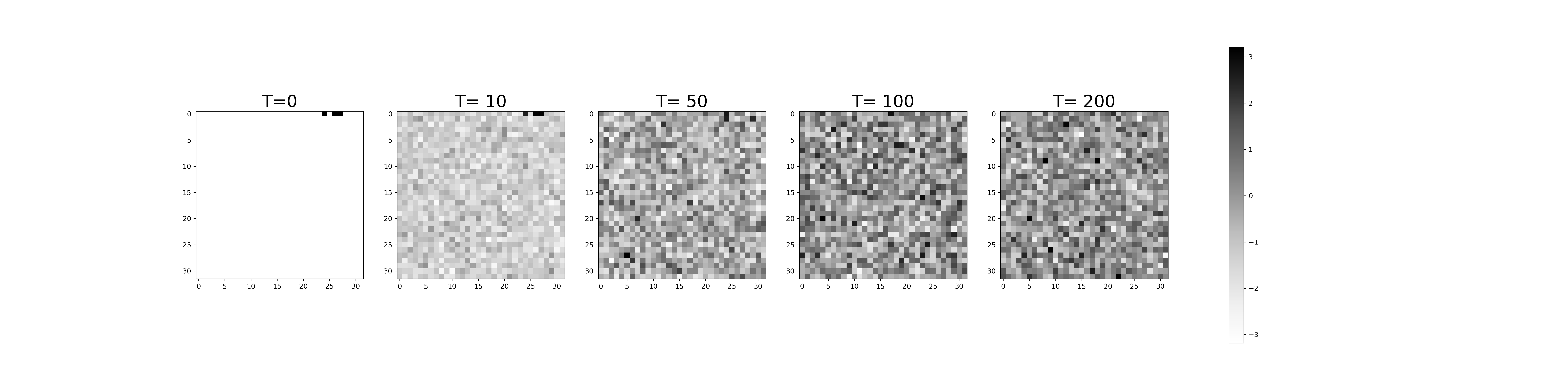}  
    \caption{Forward Process: Adding random Gaussian noise}
    \label{fig:forward}
\end{subfigure}
\begin{subfigure}{.48\textwidth}
    \centering
    \includegraphics[width=1.25\linewidth]{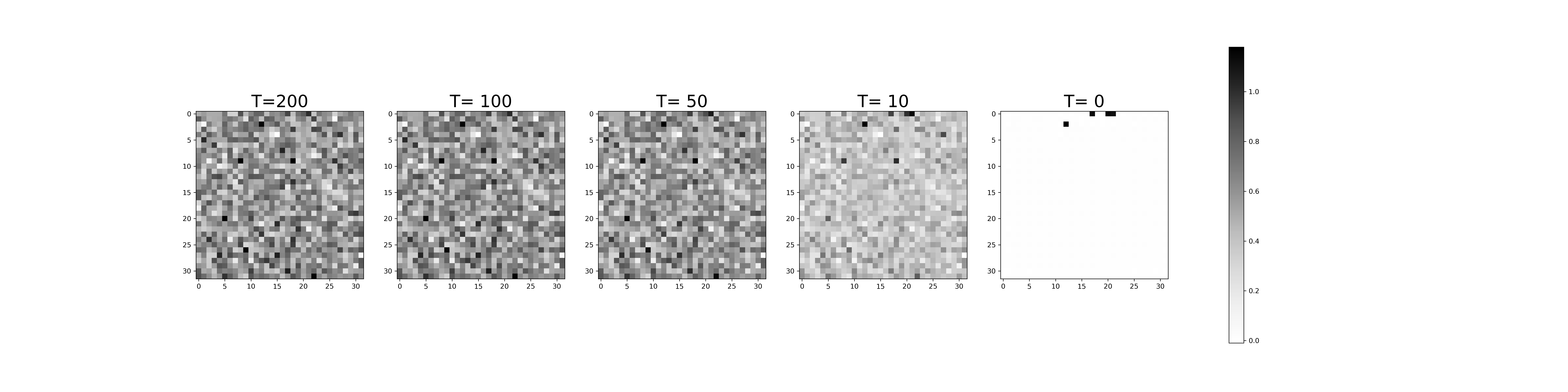}  
    \caption{Reverse Process: Intermediate samples}
    \label{fig:reverse}
\end{subfigure}
\caption{Visualization of the forward and backward process. It is worthwhile to note that \methodName~can reconstruct the noised input $X_T$ by by fixing the posterior variance $\sigma_t$ as 0 and running the denoising step.}
\label{fig:forwardandreverse}
\end{figure*}

To better understand the intermediate steps of \methodName, we visualize the resulting images at $T=0, 10, 50, 100,$ and $200$ using one sample for illustration.
Figure \ref{fig:forwardandreverse} depicts the forward and reverse process of a diffusion model. 
From Figure \ref{fig:forward} we can see that the forward process destroys the input by adding scaled random Gaussian noise step by step. At $T=200$, the generated sample looks like random noise.
If we can reverse the above process and sample from $q(x_{t-1}|x_t)$, we will be able to recreate the true sample from a Gaussian noise input. 
Since we cannot easily estimate $q(x_{t-1}|x_t)$, we learn a model $p_\theta$ using neural networks to approximate these conditional probabilities in order to run the reverse diffusion process. Figure \ref{fig:reverse} shows that given a random noise, \methodName~ is able to generate a new sample.
Furthermore, \methodName~ can reconstruct the input if it uses a non-perturbed sampling procedure (set the posterior variance $\sigma_t$ as $0$) and the noise input for the reverse process is exactly the last noisy input from a forward process. 
This confirms that by setting the posterior variance $\sigma_t$ as a nonzero number and not sharing the destroyed inputs after the forward process, \methodName~can achieve sample diversity without leaking the original inputs.

\subsection{Accelerated Generation Process}
Although the above architecture can provide satisfying results, the speed of generation process can still be a bottleneck.
The dilemma is that a small $T$ usually performs worse than a larger $T$, but a larger $T$ requires a longer generation process (and time).
To alleviate this issue, we propose an acceleration algorithm from the perspective of iterative methods.
We utilize Anderson Acceleration (AA) \cite{anderson1965iterative} to run the generation process. AA is a method that accelerates the convergence of fixed-point iterations. The idea is to approximate the final solution using a linear combination of the previous $k$ iterates. Since solving the proper combination of iterates is a nonlinear procedure, AA is also known as a nonlinear extrapolation method. 
We start with the generic truncated version of the AA prototype (see Algorithm \ref{suppalg:AA}) and explain its implementation. 
For each iteration $t \geq 0$, AA solves a least squares problem with a normalization constraint. The intuition is to minimize the norm of the weighted residuals of the previous $k$ iterates. 
\begin{algorithm}[t!]
        \SetAlgoLined
        \KwIn{Initial point $w_0$, Anderson restart dimension $k$, fixed-point mapping $g: \mathbf{R}^{n} \rightarrow \mathbf{R}^{n}.$}
        \KwOut{${w}_{t+1}$ }
        \For{t = 0, 1, \dots}{
        Set $p_t = \min\{t,k\}$.\\
        Set $F_t=[f_{t-p_t}, \dots, f_t$], where $f_i=g(w_i)-w_i$ for each $i \in [t-p_t, t]$.\\
        Determine weights $\mathbf{\beta}=(\beta_0,\dots, \beta_{p_t})^{T}$ that solves
        $
        \min _{\mathbf{\beta}}\left\|F_{t} \mathbf{\beta}\right\|_{2}, \text { s. t. } \sum_{i=0}^{p_{t}} \beta_{i}=1$.
        \\
        Set $w_{t+1}=\sum_{i=0}^{p_{t}} \beta_{i} g\left(w_{t-p_{t}+i}\right)$.
        }
         \caption{Anderson Acceleration Prototype}
         \label{suppalg:AA}
    \end{algorithm}
The constrained linear least-squares problem in Algorithm \ref{suppalg:AA} can be solved in a number of ways. Our preference is to recast it in an unconstrained form suggested in \cite{fangqr,walkerAA} that is straightforward to solve and convenient for implementing efficient updating of QR. We present the idea of the Quick QR-update Anderson Acceleration implementation as described in \cite{walkerAA} in the Supplemental Section \ref{supsec:QR}.

Here we provide the theoretical motivation and results. There is a long history of applying AA to Picard's iterative method for solving differential equations \cite{pollock2019anderson}. We can regard the updating rule of DDIM as the integrator of \eqref{eq:integrator-ode} which is approximated by $\frac{1}{2}[\bal_t \bep_{\theta}(\bx_t) - \bal_{t}^2\bx_{t}]$. Thus, \eqref{eqn:update} can be written as the Picard iteration 
\begin{equation}
    \bx_s = \bx_T + \int_{T}^{s}\underbrace{\frac{1}{2}[\bal_t \bep_{\theta}(\bx_t) - \bal_{t}^2\bx_{t}]dt}_{F(\bx_t,t)}.
\end{equation} 
Under this lens, we can apply AA to this sequence $\{\bx_{t}\}$. 

Assuming $F(\bx_t,t)$ is uniformly Lipschitz continuous in $\bx$ (a common assumption made in neural ODE literature \cite{chen2018neural}) and following the results from Theorem 2.3 in \cite{toth2015convergence}, we can derive the following acceleration guarantees for applying AA to iterates (i.e., intermediate samples) of DDIM.
\begin{theorem}
\label{thm:AA}
Assume operator $F$ has a fixed point $\bx^*$, and satisfies the following two conditions \begin{enumerate}    \item F is Lipschitz continuously differentiable in a ball $B(r) = \{\bx \in \mathbb{R}^n: \|\bx-\bx^*\|<r\}$ for some $r>0$.
    \item F is locally L-Lipschitz on $B(r)$ where $L<1$.
\end{enumerate}
Then if $\|\gamma^{t}\|_{l_1}$ is uniformly bounded by $C_{\gamma}$ for all $t>0$, \methodName~converges to $\bx^*$ locally with contraction ratio $L<\hat{L}<1$.
\end{theorem}
\begin{proof}
Here we provide a sketch of the proof. For more details, we refer readers to \cite{toth2015convergence}. Set $G(\bx) = F(\bx,t) - \bx$ and $\delta = \bx - \bx^*$. For $\hat{r}<r$ sufficiently small and $\bx \in B(\hat{r})$, it can be shown that $\|G(\bx) - G^{'}(\bx^*)\delta\|\leq \frac{\gamma}{2}\|\delta\|^2$ and $\|\delta\|(1-L)\leq \|G(\bx)\| \leq (1+L)\|\delta\|$ where $\gamma$ is the Lipschitz constant of $G^{'}(\bx)$ on $B(\hat{r})$. Then we have $\|G(\bx_t)\|\leq \hat{L}^{k}\|G(\bx_0)\|$ for all $0\leq t\leq T$, which obviously holds for $T=0$. Leveraging the two inequalities above, it can be shown that
\begin{equation*}
    \|G(\bx_{T+1})\|\leq \underbrace{\frac{\frac{L}{\hat{L}} + \Big(\frac{C_{\alpha}\gamma \hat{r}}{2(1-L)}\Big)\hat{L}^{-k-1}}{1-\frac{\gamma r}{2(1-L)}}}_{C\leq 1}\hat{L}^{T+1}\|G(\bx_0)\|
\end{equation*}
To ensure $C\leq 1$, just reduce $\hat{r}$ until it satisfies $\hat{r}<\frac{2(1-L)}{\gamma}$.
\end{proof}

Theorem \ref{thm:AA} provides the theoretical justification for leveraging AA to predict the next denoised intermediate sample. Since \methodName~requires much fewer iterations and reduced computational costs to generate realistic samples, it can dramatically save inference time.

\subsection{Conditioned Sampling}
Sometimes it is insufficient for the model to produce realistic-looking data, it should also ensure the generated examples preserve utility in a down-stream task.
As such, if a particular class label is passed to the synthesizer, it should produce a health record that matches the distribution of that label.
This is one of the limitations of MedGAN and CorGAN, in that it may not preserve the class-dependent label information.
We equip \methodName~with this ability by incorporating the idea of a classifier-guided sampling process \cite{dhariwal2021diffusion}. From a conceptual level, the estimated noise ${\epsilon_\theta}(\bx_t, t)$ in each step is deducted by $\sqrt{1-\bar{\alpha}_{t}} \nabla_{\bx_{t}} \log f_{\phi}\left(y \mid \bx_{t}\right)$, where $f_{\phi}\left(y \mid \bx_{t}\right)$ is a trained classifier on the noisy $\bx_t$.
This modified updating rule tends to up-weighting the probability of data where the classifier $f_{\phi}\left(y \mid \bx_{t}\right)$ assigns high likelihood to the correct label.
Algorithm \ref{suppalg:AASampling-dff} summaries the corresponding accelerated and conditioned sampling
algorithm.
\begin{algorithm}[t!]
      Choose a small integer $k$ as Anderson restart dimension\\
      Given a random noise $\bx_{T}\sim\stdnormal$ and label $y$\\
        \For{$t = T,\dots,1$}
        {
        $\bz \sim \stdnormal$ if $t>1$, else $\bz =0$\\
        $\epsilon= \epsilon_{\theta}\left(\bx_{t}\right)-\sqrt{1-\bar{\alpha}_{t}} \nabla_{\bx_{t}} \log p_{\phi}\left(y \mid \bx_{t}\right)$\\
        Update $\bx_{t-1}$ according to equation \eqref{eqn:update}\\
        $\bx_{t-1} = Anderson(\bx_t, \bx_{t-1}, k)$
        }
        \Return{$\bx_0$}
         \caption{Accelerated Conditioned Sampling}
         \label{suppalg:AASampling-dff}
    \end{algorithm}

\section{Experiments}
In this section, we answer following questions:
\textbf{Q1:} Can \methodName~generate high quality synthetic EHRs compared to existing methods?
\textbf{Q2:} Can \methodName~accurately generate conditioned samples that match the distribution of the given label?
\textbf{Q3:} Can our proposed acceleration technique alleviate the issue of slow generation process of a diffusion model? How is its performance under different settings? 
We first describe the two health datasets used in our experiments. Next, we give an overview of baseline methods and implementation details. We then evaluate the effectiveness of the proposed acceleration technique with an ablation study. 

\subsection{Datasets}
We use the following two publicly available datasets.
\begin{enumerate}
    \item MIMIC-III \cite{mimic}: A large database containing de-identified health data associated with approximately sixty thousand admissions of critical care unit patients from the Beth Israel Deaconess Medical Center collected between 2001 and 2012. For each patient, we extract the International Classification of Diseases (ICD-9) diagnosis codes. We represent a patient record as a fixed-size vector with 1071 entries for each patient record. The pre-processed dataset is a 46520 $\times$ 1071 binary matrix and used to evaluate the binary discrete variable generation and the proposed acceleration technique.
    \item Patient Treatment Classification\footnote{\url{https://www.kaggle.com/manishkc06/patient-treatment-classification}}: A dataset collected from a private hospital in Indonesia. It contains the 8 different laboratory test results of 3309 patient used to determine next patient treatment whether in care or out care. We use this datasets to perform continuous synthetic EHR generation and investigate the effectiveness of conditional generation of \methodName.
\end{enumerate}

\begin{figure*}
\centering
\begin{subfigure}{.24\textwidth}
    \centering
    \includegraphics[width=.95\linewidth]{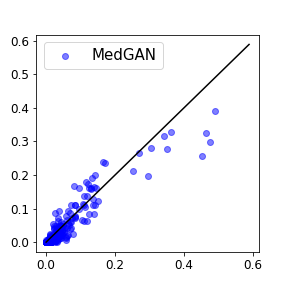}  
    \caption{$\rho=0.94, SAE=5.68$}
    \label{SUBFIGURE LABEL 1}
\end{subfigure}
\begin{subfigure}{.24\textwidth}
    \centering
    \includegraphics[width=.95\linewidth]{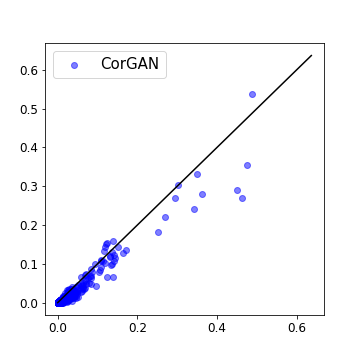}  
    \caption{$\rho=0.97, SAE=4.49$}
    \label{SUBFIGURE LABEL 2}
\end{subfigure}
\begin{subfigure}{.24\textwidth}
    \centering
    \includegraphics[width=.95\linewidth]{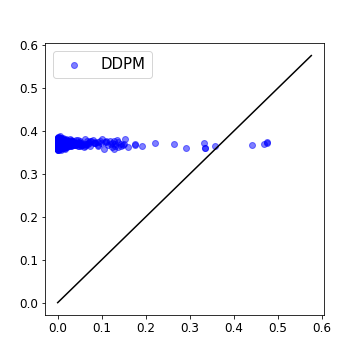}  
    \caption{$\rho=-0.06, SAE=366$}
    \label{SUBFIGURE LABEL 3}
\end{subfigure}
\begin{subfigure}{.24\textwidth}
    \centering
    \includegraphics[width=.95\linewidth]{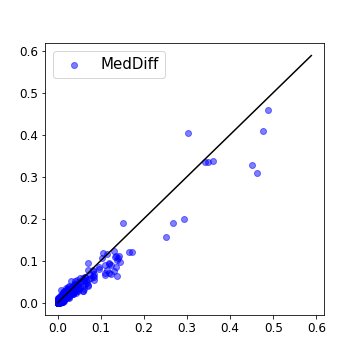}  
    \caption{$\rho=0.98, SAE=4.16$}
    \label{SUBFIGURE LABEL 4}
\end{subfigure}
\caption{The scatter plots of dimension-wise probability. Each point depicts one unique diagnosis code. The x-axis and
y-axis represent the Bernoulli success probability for real and synthetic datasets, respectively. The diagonal line shows the ideal case.}
\label{fig:correlation}
\end{figure*}

\subsection{Baselines and Implementation}
We compare \methodName~ with following methods.
\begin{enumerate}
    \item MedGAN \cite{pmlr-v68-choi17a} \footnote{Code available at \url{https://github.com/astorfi/cor-gan}}: A GAN-based model that generates the low-dimensional synthetic records and decodes them with an autoencoder.
    \item CorGAN \cite{Torfi2020CorGANCC}$^2$: A GAN-based model similar to MedGAN but combines Convolutional Generative Adversarial Networks and Convolutional Autoencoders.   
    \item Noise Conditional Score Network (NCSN) \cite{score}\footnote{Code adopted from \url{https://github.com/acids-ircam/diffusion_models}.}: Instead of directly learning the probability of the data log $p(x)$, this method aims to learn the gradients of log $p(x)$ with respect to $x$. This can be understood as learning the direction of highest probability at each point in the input space. 
    After training, the sampling process is achieved by applying Langevin dynamics.
    \item DDPM \cite{ho2020denoising}\footnote{Code adopted from \url{https://github.com/lucidrains/denoising-diffusion-pytorch}}: A class of latent variable models inspired by considerations from nonequilibrium thermodynamics \cite{denoise}. 
\end{enumerate}
We implemented \methodName~with Pytorch. For training the models, we used Adam \cite{kingma2017adam} with the learning rate set to $0.001$, and a mini-batch of $128$ and $64$ for MIMIC-III and the patient treatment dataset, respectively on a machine equipped with one Nvidia GeForce RTX 3090 and CUDA 11.2. Hyperparamters of \methodName~are selected after grid search. We use a timestep $T$ of 200, a noise scheduling $\beta$ from $1 \times 10^{-4}$ to $1 \times 10^{-2}$ and a table size $k=3$. The code will be publicly available upon publication.

\subsection{Sample Quality Evaluation}
We first evaluate the effectiveness of \methodName~on MIMIC-III. Following previous works CorGAN and MedGAN, we use the dimension-wise probability as a basic evaluation metric to
determine if \methodName~can learn the distribution of the real data (for each dimension). This measurement refers to the Bernoulli success probability of each dimension (each dimension is a unique ICD-9 code). We report the dimension-wise probability in Figure \ref{fig:correlation}. We also use the correlation coefficient $\rho$ and sum of absolute errors (SAE) as our quantitative metrics. From Figure \ref{fig:correlation}, we can observe \methodName~shows the best performance both in terms of highest correlation ($\rho=0.98)$ and the lowest SAE (4.16). The results also illustrate that na{\"i}ve application of DDPM yields substandard results, even worse than MedGAN and CorGAN.

\subsection{Conditioned Sample Quality Evaluation}
\begin{figure*}
\centering
\begin{subfigure}[b]{\textwidth}
\centering
   \includegraphics[width=0.83\linewidth]{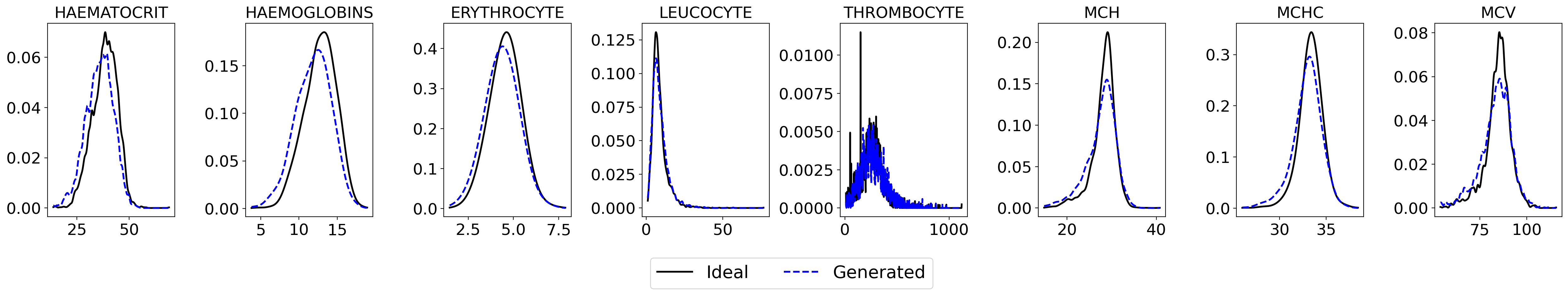}
   \caption{\textit{Black:} KDE for all 3309 patient records. \textit{Blue:} KDE for 3309 Unconditioned synthetic records. }
   \label{fig:kdeall} 
\end{subfigure}
\begin{subfigure}[b]{\textwidth}
\centering
   \includegraphics[width=0.83\linewidth]{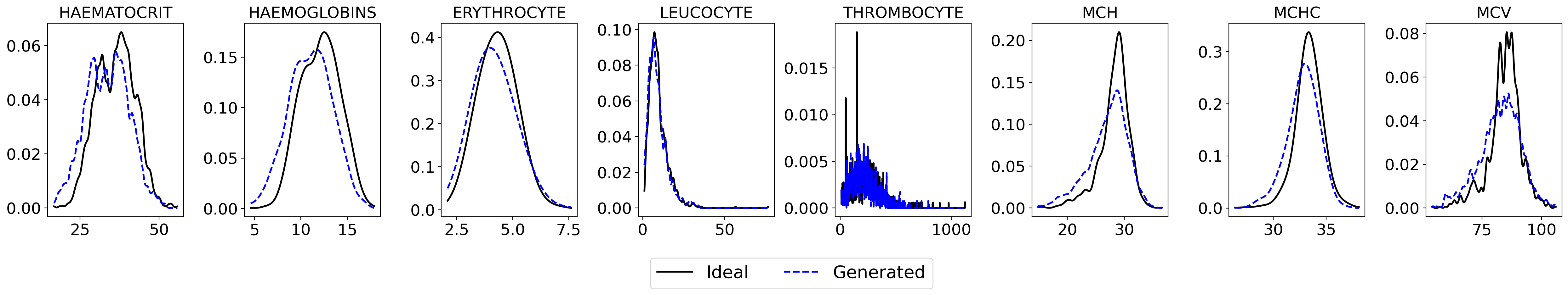}
   \caption{\textit{Black:} KDE for 1317 in-care patient records. \textit{Blue:} KDE for 1317 conditioned synthetic in-care patient records.}
   \label{fig:kdein} 
\end{subfigure}
\begin{subfigure}[b]{\textwidth}
\centering
   \includegraphics[width=0.83\linewidth]{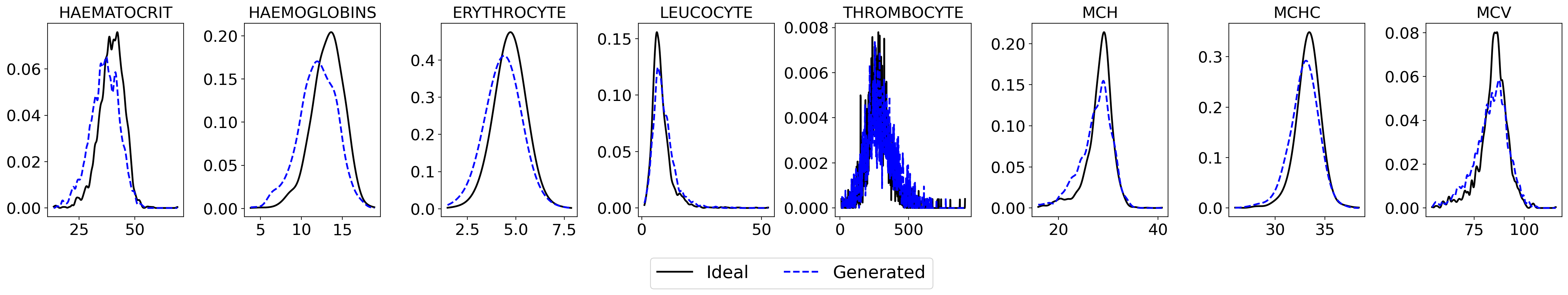}
   \caption{\textit{Black:} KDE for 1992 out-care patient records. \textit{Blue:} KDE for 1992 conditioned synthetic out-care patient records.}
   \label{fig:kdeout}
\end{subfigure}
\caption{Kernel density estimation (KDE) for each feature. \textit{Black} is the true density. \textit{Blue} is the generated samples density.
}
\end{figure*}

Next, we answer the question whether \methodName~preserves the conditioned sampling of health records with a given label on the Patient Treatment Classification dataset. We compare the underlying probability density function of the original dataset and generated samples using Gaussian Kernel Density Estimation (KDE). We note that none of the baseline methods offer this capability.
The results for the unconditional sampling are shown in Figure \ref{fig:kdeall} while the conditional sampling versions are shown in Figures \ref{fig:kdein} and \ref{fig:kdeout}.
From the plots, the generated synthetic samples mostly follow the overall probability function of the original dataset.
Furthermore, we can observe that the distribution of in-care patients and out-care patients are different in terms of shape, local modes, and range. This means the generated samples are potentially more informative and offer more utility when used for data sharing. This experiment also demonstrates the flexibility and advantage of \methodName~in terms of ability to deal with class-conditional sampling.

\subsection{Effects of Accelerated Sampling}
\begin{figure*}
\centering

\begin{subfigure}[b]{0.48\textwidth}
\centering
   \includegraphics[width=1\linewidth]{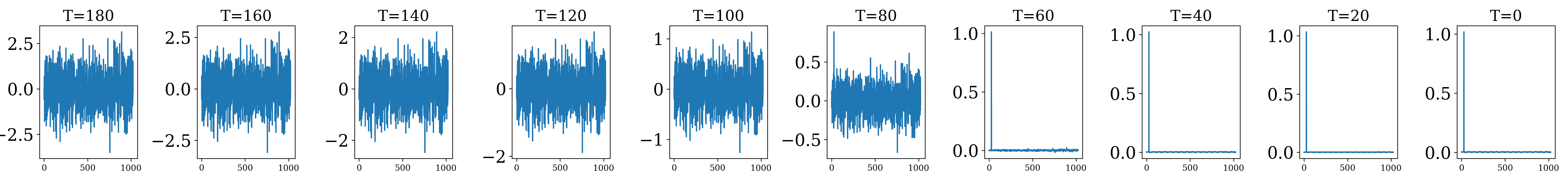}
   \caption{Regular Reverse Process for $T=200$: it takes about 160 iterations to converge. }
   \label{fig:T200reg} 
\end{subfigure}
\begin{subfigure}[b]{0.48\textwidth}
\centering
   \includegraphics[width=1\linewidth]{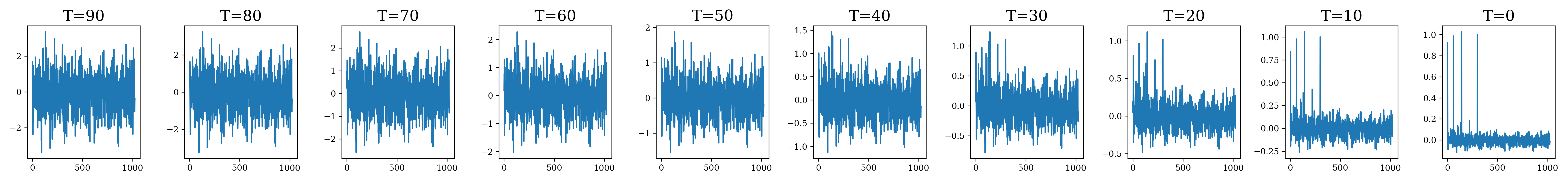}
   \caption{Regular Reverse Process for $T=100$: the generated sample is still very noisy after 100 iterations.}
   \label{fig:T100reg} 
\end{subfigure}
\begin{subfigure}[b]{0.48\textwidth}
\centering
   \includegraphics[width=1\linewidth]{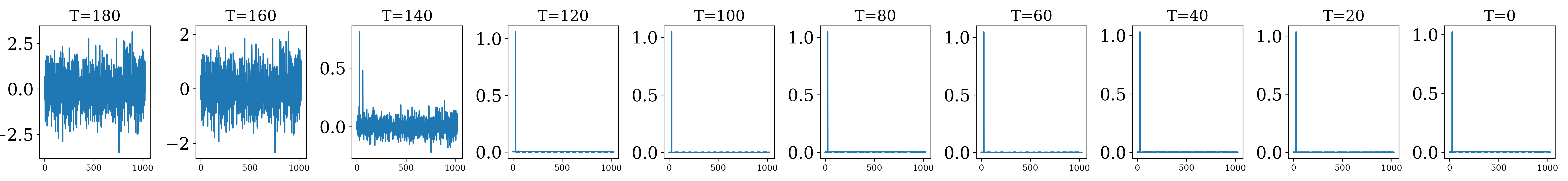}
   \caption{Accelerated Reverse Process for $T=200$: it takes about 80 iterations to converge, which implies a $2\times$ speed up. }
   \label{fig:T200AA} 
\end{subfigure}
\begin{subfigure}[b]{0.48\textwidth}
\centering
   \includegraphics[width=1\linewidth]{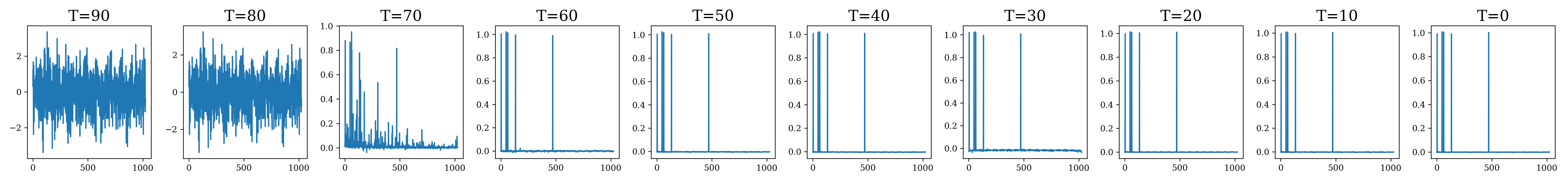}
   \caption{Accelerated Reverse Process for $T=100$: After 40 iteration, \methodName~ is able to generate a high-quality sample.}
   \label{fig:T100AA}
\end{subfigure}
\caption[Two numerical solutions]{Evolution plot of regular generation process (\ref{fig:T200reg}, \ref{fig:T100reg}) and \methodName~(\ref{fig:T200AA}, \ref{fig:T100AA}). }
\end{figure*}

Next, we evaluate the usefulness of our proposed accelerated sampling algorithm on MIMIC-III. For $T=200$, we present the evolution process of generating one sample by use of the regular procedure in Figure \ref{fig:T200reg} and \methodName~in Figure \ref{fig:T200AA}. Similarly, Figures \ref{fig:T100reg} and \ref{fig:T100AA} depict the evolution process of generating one sample for $T=100$. 
The iteration number is defined as $T-t$. When $T=200$, we can observe \methodName~converges  $\sim80$ iterations whereas the regular sampling process takes $\sim160$ iterations, a $2\times$ speed up. 
From Figures \ref{fig:T100AA} and \ref{fig:T100reg}, \methodName~can generate a high-quality sample when the regular sampling process fails to converge.  

\begin{figure*}
\centering
\begin{subfigure}{.24\textwidth}
    \centering
    \includegraphics[width=.95\linewidth]{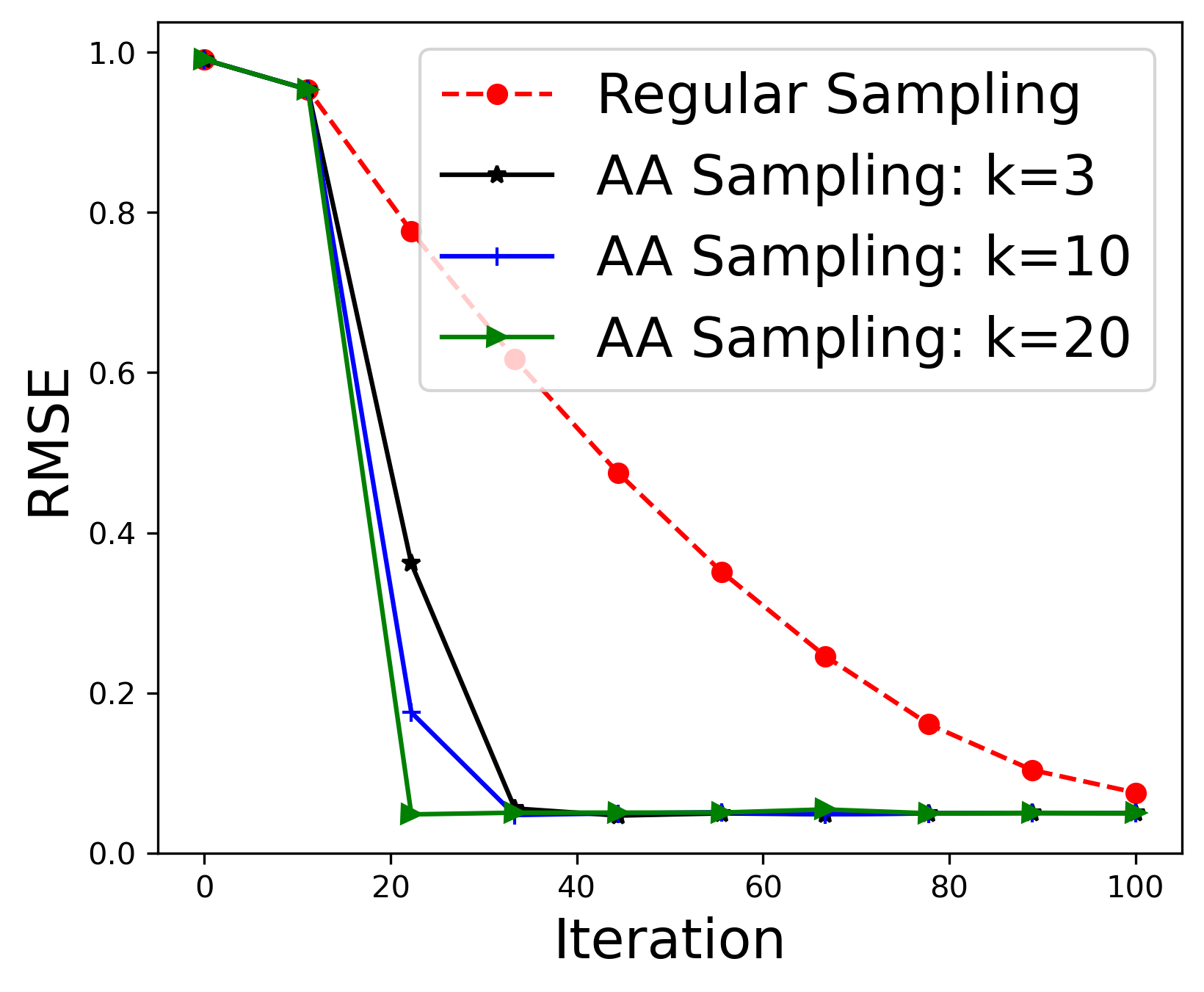}  
    \caption{Comparison in terms of iterations, $T=100$}
    \label{fig:t100iter}
\end{subfigure}
\begin{subfigure}{.24\textwidth}
    \centering
    \includegraphics[width=.95\linewidth]{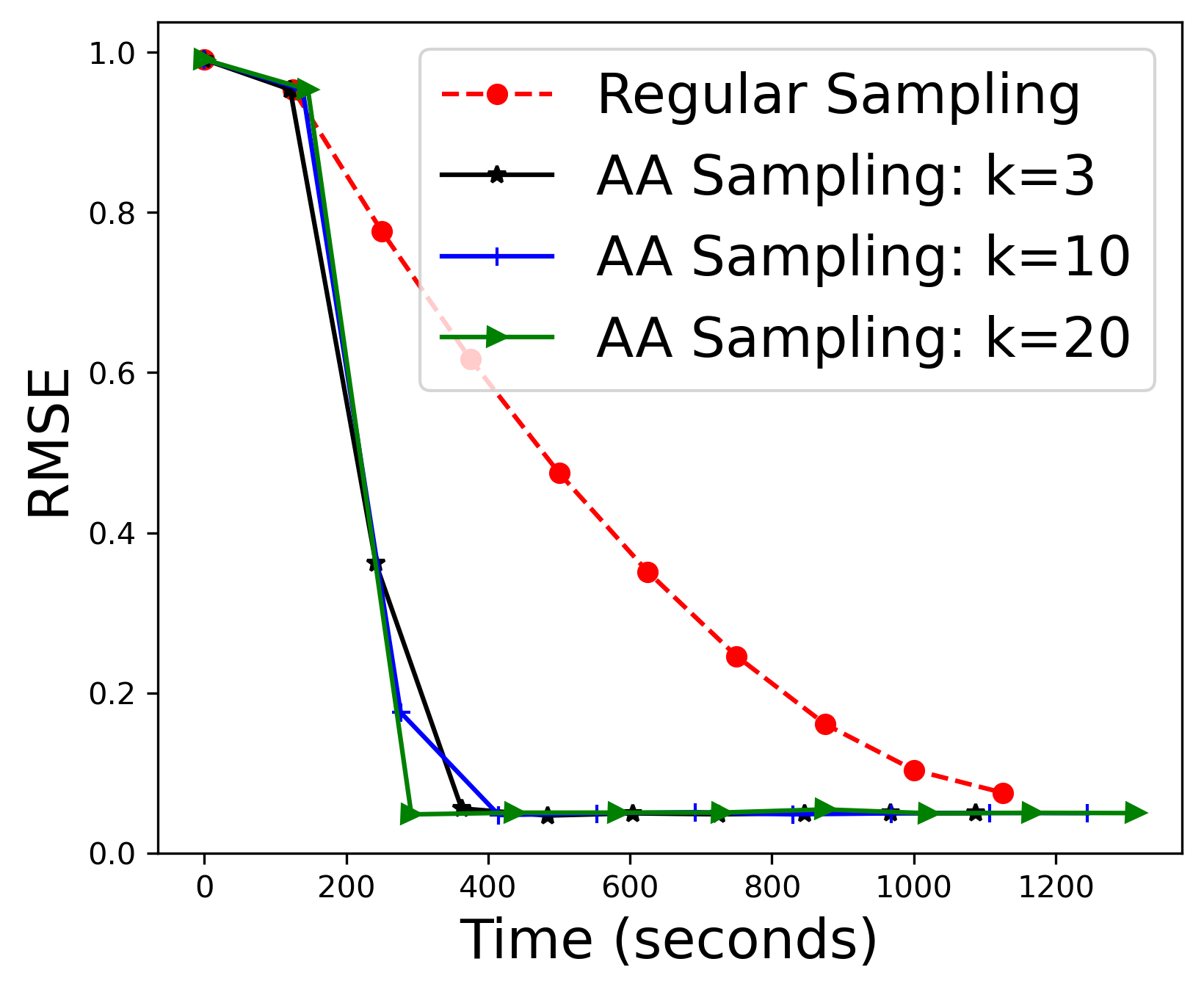}  
    \caption{Comparison in terms of time, $T=100$}
    \label{fig:t100time}
\end{subfigure}
\begin{subfigure}{.24\textwidth}
    \centering
    \includegraphics[width=.95\linewidth]{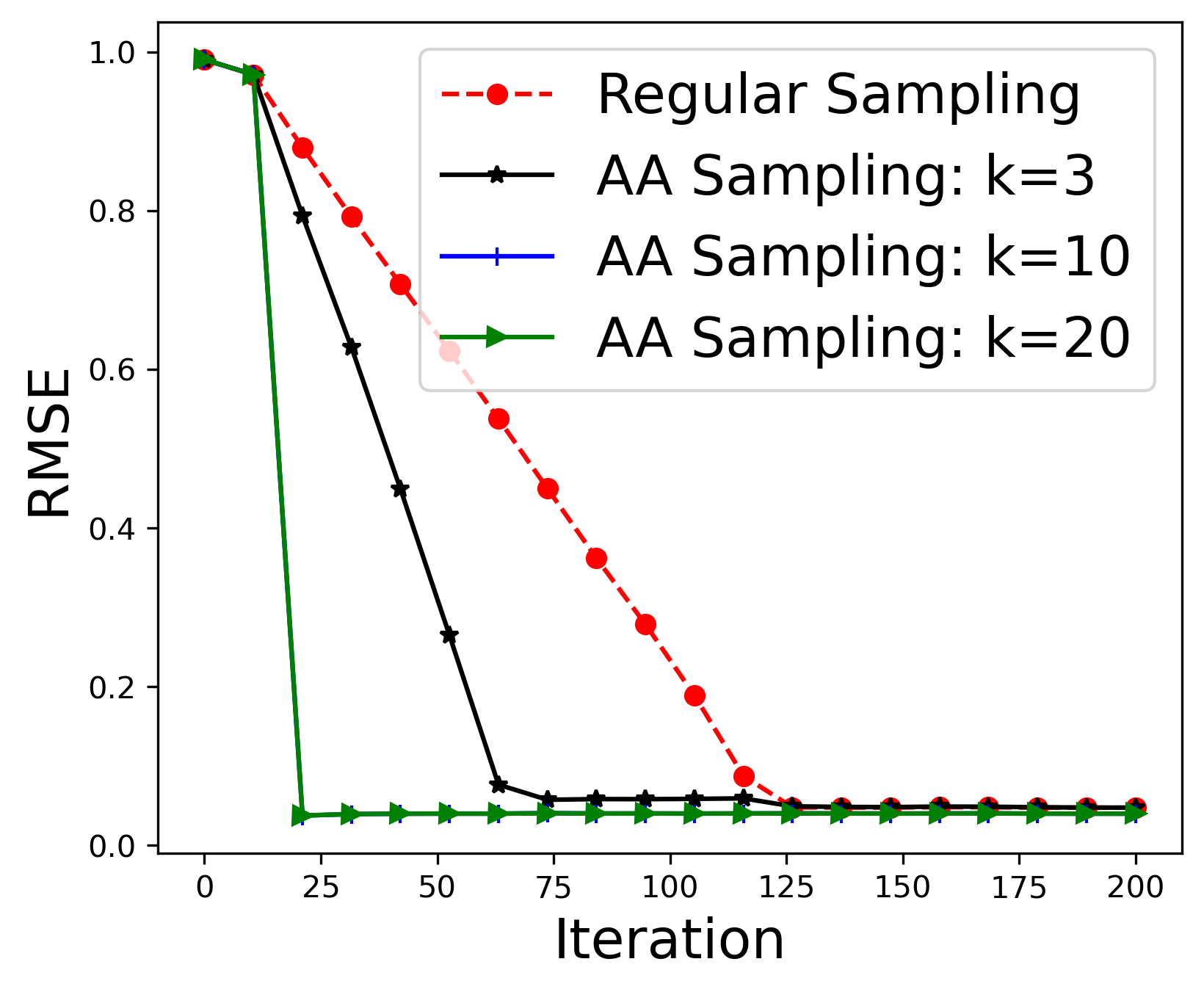}  
    \caption{Comparison in terms of iterations, $T=200$}
    \label{fig:t200iter}
\end{subfigure}
\begin{subfigure}{.24\textwidth}
    \centering
    \includegraphics[width=.95\linewidth]{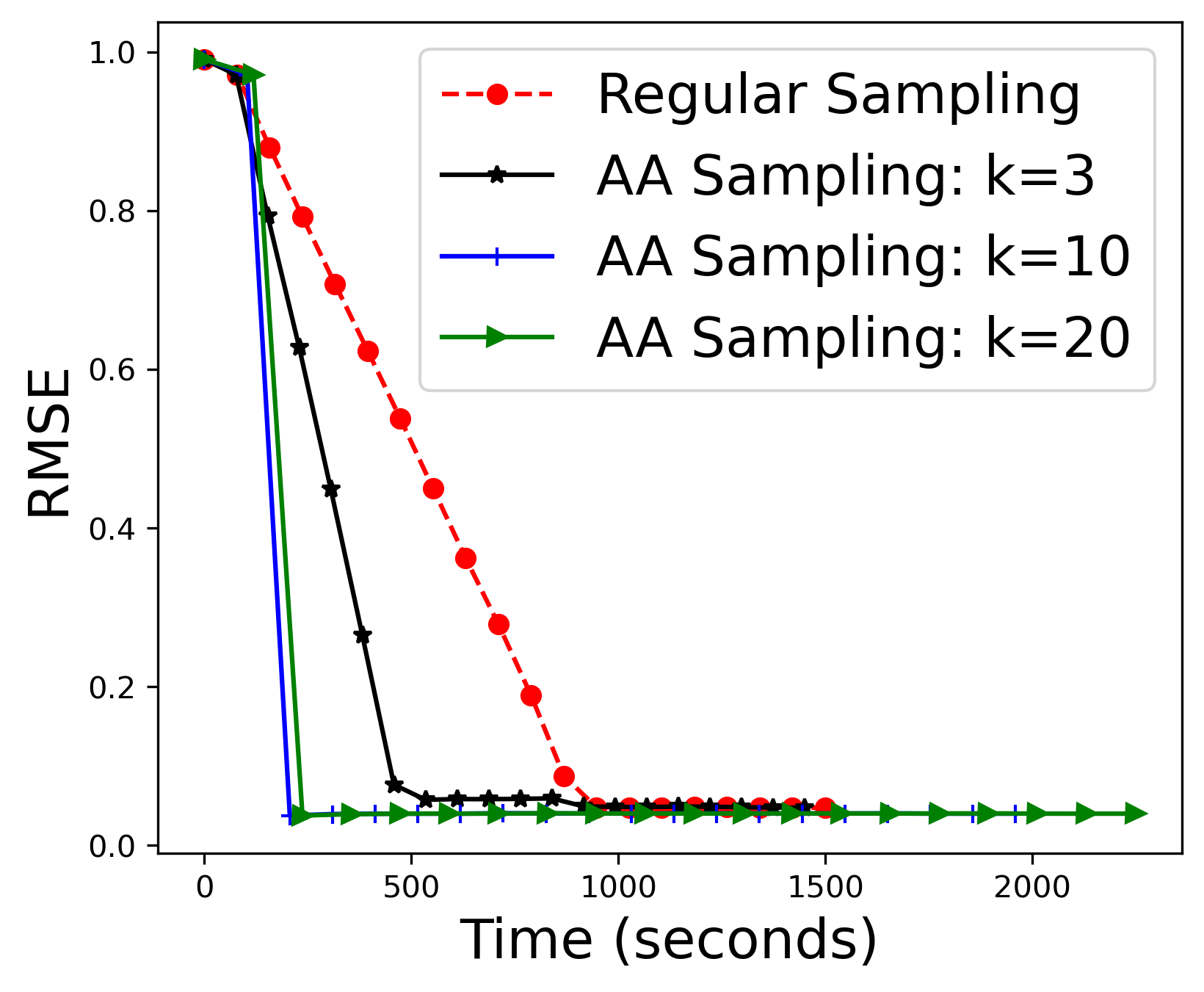}  
    \caption{Comparison in terms of time, $T=200$}
    \label{fig:t200time}
\end{subfigure}
\caption{Results of accelerated sampling versus regular sampling with different $T$ and $k$.}
\label{fig:ablationAA}
\end{figure*}

We perform an additional qualitative ablation study of \methodName~to investigate whether our model has actually accelerated the generation process with different hyperparameters. We use the root mean squared error of the Bernoulli success probability of each dimension for the generated samples and original dataset, which is defined as follows,
\begin{equation}
RMSE = \sqrt{\frac{\sum_{d=1}^{D}\left(p_{d}-\hat{p}_{d}\right)^{2}}{D}},
\end{equation}
where $D$ is the feature size. We run \methodName~using different $T$ and AA restart dimension (or table size) $k$ and plot the results in Figure \ref{fig:ablationAA}. Figures \ref{fig:t100iter} and \ref{fig:t200iter} indicate an increasing of table size results in faster convergence in terms of iteration
number. Figures \ref{fig:t100time} and \ref{fig:t200time} show that although there is additional computation, it does not hinder the benefits of adopting the accelerated algorithm with a small $k$. Moreover, Figures \ref{fig:t100iter} and \ref{fig:t200iter} verify that adopting the accelerated algorithm can help to generate high-quality samples if the model is trained with fewer timesteps.

\subsection{Case Study: Discriminative Analysis}
Classification models are often developed on EHR data to determine whether the next patient treatment is in-care or out-care. To evaluate the utility of our synthetic EHRs, we evaluate on a prediction task. We use an 80/20 training/test split on the patient treatment dataset. We use the training set (2647 patients) to train and generate class-conditional synthetic data. Then, we evaluate the performance of a logistic regression model trained using the real data or the synthetic data on the real test set (662 patients). Unsurprisingly, there was a drop in AUC between the real data (0.766) and the synthetic data (0.742).

\begin{figure}
\centering
\includegraphics[width=0.6\linewidth]{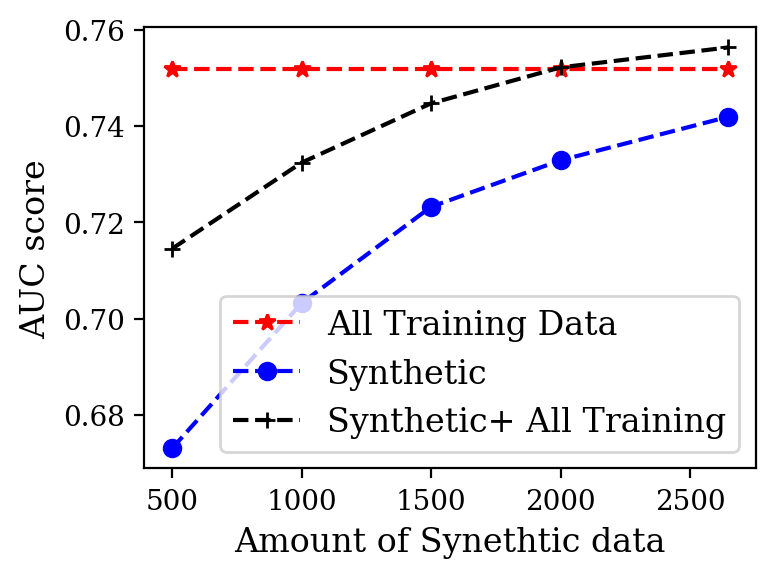}  
\caption{Data augmentation performance of \methodName~as a function of the number of synthetic records.}
\label{fig:dataaug}
\end{figure}

Synthetic EHRs can also be beneficial for data augmentation purposes to develop a more robust classifier. We also investigate whether the data generated by \methodName~can be used for this purpose by varying the amounts of synthetically generated data used to augment the training set, and evaluate the AUC on the test set.
The results from this experiment are shown in Figure \ref{fig:dataaug}. It can be observed that augmenting the training set with more than 2000 synthetic records can yield better performance than just the real data.
However, augmenting with small amounts of synthetic data can harm the performance as shown by the performance degradation with less than 2000 synthetic records. We posit that this might be that the synthetic sample distribution may not yet reflect the true distribution with insufficient samples.

\section*{Acknowledgements}
This work was supported by the National Science Foundation awards IIS-1838200,  IIS-2145411, and DMS-2208412. Any opinions, findings, conclusions or recommendations expressed in this material are those of the authors and do not necessarily reflect the views of the funders. The authors declare that there are no conflict of interests.

\bibliographystyle{siam}
\bibliography{main}

\clearpage


\section{Supplemental Material}
\subsection{Fast and Efficient Implementation of Anderson Acceleration}
\label{supsec:QR}
Define $f_i=g(w_i)-w_i$, $\triangle f_i= f_{i+1}-f_i$ for each $i$ and set $F_t=[f_{t-p_t}, \dots, f_t$], $\mathcal{F}_t=[\triangle f_{t-p_t}, \dots, \triangle f_t]$. Then solving the
least-squares problem ($\min _{\mathbf{\beta}}\left\|F_{t} \mathbf{\beta}\right\|_{2}, \text { s. t. } \sum_{i=0}^{p_{t}} \beta_{i}=1$) is equivalent to 
\begin{equation}
\label{eqn:unAA}
\min _{\gamma=\left(\gamma_{0}, \ldots, \gamma_{p_{t-1}}\right)^{T}}\left\|f_{t}-\mathcal{F}_{t} \gamma\right\|_{2}
\end{equation}
where $\alpha$ and $\gamma$ are related by $\alpha_{0}=\gamma_{0}, \alpha_{i}=\gamma_{i}-\gamma_{i-1}$ for $1 \leq i \leq p_{t}-1$, and $\alpha_{p_{t}}=1-\gamma_{p_{t}-1}$.

Now the inner minimization subproblem can be efficiently solved as an unconstrained least squares problem by a
simple variable elimination. This unconstrained least-squares problem leads to a modified form of Anderson acceleration
\begin{equation}
\begin{aligned}
    w_{t+1} &= g\left(w_{t}\right)-\sum_{i=0}^{p_{t}-1} \gamma_{i}^{(t)}\left[g\left(w_{t-p_{t}+i+1}\right)-g\left(w_{t-p_{t}+i}\right)\right] \\
    &= g\left(w_{t}\right)-\mathcal{G}_{t} \gamma^{(t)}
\end{aligned}
\end{equation}
where $\mathcal{G}_t=[\triangle g_{t-p_t}, \dots, \triangle g_{t-1}]$ with  $ \triangle g_i= g(w_{i+1})-g(w_i)$ for each $i$. 

To obtain $\gamma^{(t)}=\left(\gamma_{0}^{(t)}, \ldots, \gamma_{p_{t}-1}^{(t)}\right)^{T}$ by solving \eqref{eqn:unAA} efficiently, we show how the successive least-squares problems can be solved efficiently by updating the factors in the QR decomposition $\mathcal{F}_t = Q_tR_t$ as the algorithm proceeds. We assume a thin QR decomposition, for which the solution of the least-squares problem is obtained by solving the $p_t \times p_t$ linear system $R \gamma=Q^{\prime}*f_{t}$. Each $\mathcal{F}_t$ is $n \times p_t$ and is obtained from $\mathcal{F}_{t-1}$ by adding a column on
the right and, if the resulting number of columns is greater than $p$, also cleaning up (re-initialize) the table. That is,we never need to delete the left column because cleaning up the table stands for a restarted version of AA. As a result, we only need to handle two cases; 1 the table is empty (cleaned). 2 the table is not full.
When the table is empty, we initialize $\mathcal{F}_1=Q_1R_1$ with $Q_1=\triangle f_{0} /\left\|\triangle f_{0}\right\|_{2}$ and $R=\left\|\triangle f_{0}\right\|_{2}$. If the table size is smaller than $p$,  we add a column on the right of $\mathcal{F}_{t-1}$. Have $\mathcal{F}_{t-1} =QR$, we update $Q$ and $R$ so that $\mathcal{F}_{t}=\left[\mathcal{F}_{t-1}, \Delta f_{t-1}\right]=Q R$. 
It is a single modified Gram–Schmidt sweep that is described as follows:
\begin{algorithm}
        \SetAlgoLined
        \For{$i = 1,\dots, p_{t-1}$}{
        Set $R(i, p_t) = Q(:,i)' * \triangle f_{t-1}$.\\
        Update $\triangle f_{t-1}\leftarrow \Delta f_{t-1}-R\left(i, p_{t}\right) * Q(:, i)$ 
        }
        Set $Q\left(:, p_{t}\right)=\triangle f_{t-1} /\left\|\triangle f_{t-1}\right\|_{2}$ and $R\left(p_{t}, p_{t}\right)=\left\|\Delta f_{t-1}\right\|_{2}$
         \caption{QR-updating procedures}
         \label{suppalg:QR}
    \end{algorithm}
Note that we do not explicitly conduct QR decomposition in each iteration, instead we update the factors ($O(k^2n)$) and then solve a linear system using back substitution which has a complexity of $O(k^2)$. Based on this complexity analysis, we can find Anderson acceleration with QR-updating scheme has limited computational overhead. 

\subsection{Additional results}
\begin{figure*}
\centering
\includegraphics[width=1\linewidth]{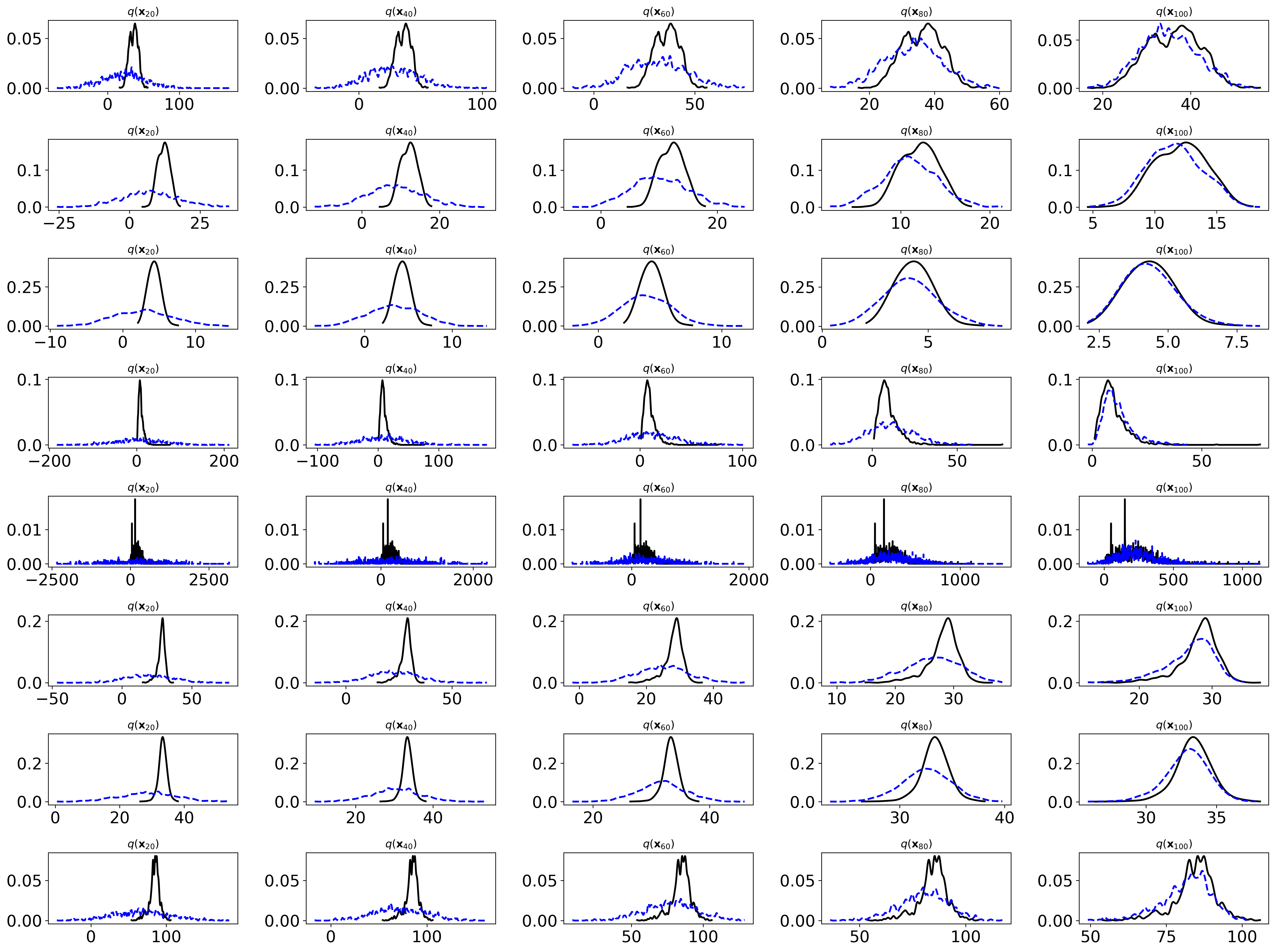}
\caption{Conditioned Sampling Process for in care patient records. Each row represents the kernel density estimation for each feature. We can observe that given a random Gaussian noise, \methodName is able to transform to the target distribution accurately. }
\label{fig:kdeouteve}
\end{figure*}

We present the evolution of the reverse (sampling) process in Figure \ref{fig:kdeouteve}. 



\end{document}